\documentclass[copyright,creativecommons]{eptcs}
\providecommand{\event}{MOD* 2014} 
\usepackage{breakurl}             

\title{An Approach to Model Checking of Multi-agent Data Analysis}
\author{Natalia Garanina
\and Eugene Bodin \institute{A.P. Ershov Institute of Informatics
Systems, Novosibirsk, Russia\thanks{The research has been
supported by Russian Foundation for Basic Research (grant
13-01-00643) and Siberian Branch of Russian Academy of Science
(Integration Grant n.15/10 ``Mathematical and
Methodological Aspects of Intellectual Information Systems'').}}\\
\email{garanina,bodin,lena@iis.nsk.su} \and Elena Sidorova}
\def\titlerunning{SPIN for Multi-agent Data Analysis}
\def\authorrunning{N. Garanina, E. Bodin, E. Sidorova}

\newtheorem{proposition}{Proposition}

\begin{document}
\maketitle

\begin{abstract}
The paper presents an approach to verification of a multi-agent
data analysis algorithm. We base correct simulation of the
multi-agent system by a finite integer model. For verification we
use model checking tool SPIN. Protocols of agents are written in
Promela language and properties of the multi-agent data analysis
system are expressed in logic LTL. We run several experiments with
SPIN and the model.
\end{abstract}

\section{Introduction}

The purpose of the paper is to apply formal verification methods
to multi-agent algorithms of data analysis in a framework of
ontology population.

Multi-agent data analysis for ontology population is a multilevel
process. Let us have an ontology, whose elements are classes,
specified by a set of (key) attributes, and relations, specified
by attributes and a set of classes they connect. Ontology
population rules depend on given ontology classes and relations.
Besides, we have rules for input data processing. These data can
be natural language text or special format of data storing, for
example, various databases or tagged internet pages. We consider
all these rules defined formally such that every rule can (1) use
data, which can be values of attributes or instances of classes or
relations; (2) bind a tuple of attributes into an instance of a
given class; (3) determine attribute values of the relation and
whether some class instances belong to a given relation.

At the first stage of multi-agent data analysis, preliminary
investigation of input data generates underdetermined objects that
can be instances of classes and relations of the predefined
ontology. At the next stage, using rules of ontology population
and data processing, concerning semantic and syntactic
consistency, these objects are evaluated from input data as full
as it is possible. At the third stage, these objects-instances
resolve ambiguities that are an inherent feature of automatic data
analysis.

At the second stage of analysis \emph{information (instance and
relation) agents} appears. They cor\-res\-pond to instances of
classes and relations. Information agents interact with \emph{rule
agents} that implement given rules of data processing and ontology
population. These agents exchange information necessary for
specification of information agents. A special \emph{controller
agent} detects system termination, i.e. a moment when all possible
information is retrieved from data and agents just waiting for
messages from each others. In contrast to all other model agents,
this service agent is universal, i.e. it does not depend on a
given ontology and input data types.

All agents act in parallel hence we have to verify some important
properties of the system. In particulary, properties to be
verified are correctness of termination detection and simple
operability of the analysis system. For checking these properties
we choose model checking tool SPIN \cite{Holzmann03}. SPIN has
rather expressive input language for specification our data
analysis model and its properties, and a well-developed system for
error detection and examination.

A multi-agent approach for information retrieval from
heterogeneous data sources for completing ontology is widespread.
In particular, it is used for natural language processing
\cite{Aref03,Ariadne98,Trojhan06,Fum88} and web processing
\cite{BanaresAlcantaraJimenezAldea05,ChengXieYang08,ClarkLazarou97}.
Agents in these works have different behaviors. Usually in web
processing, agents are high-level entities that manage rather data
flows, using standard algorithm for knowledge retrieval, than data
itself. In natural language processing, agents are either
associated with conventional linguistic levels (morphological,
syntactic, semantic) or targeted to recognize specific linguistic
phenomena such as ellipsis, anaphora, parataxis, homonymy. These
agents do not use ontological knowledge substantially. Thus they
are computing processes which may speed up information retrieval
due to their parallel work but they do not affect the retrieval
qualitatively.

Our approach implements multi-agent low-level data analysis in
which agents do not process input data by traditional methods but
present information items themselves. To the best of our knowledge
a similar approach is introduced in
\cite{MinakovRzevskiSkobelevVolman07} only. Verification of such
system is also unknown to us.

The rest of the paper is organized as follows. The next section
\ref{AgentModel} describes agents of our systems and their action
protocols. The section \ref{APPR} grounds an approach to finite
state model checking of our multi-agent model. The following
section \ref{SPIN} presents the method for expressing the
multi-agent model and its properties in SPIN. Finally, we conclude
in the last section \ref{CONC} with a discussion of further
research.

\section{Agent Model and Protocols}\label{AgentModel}

Outline of the approach and multi-agent system follows. There is
an ontology of a subject domain, a set of rules for completing it,
a semantic and syntactic model of the input data language and a
finite data to extract information for the ontology. We consider a
\emph{subject domain's ontology} to be the following tuple $O =
\langle C, R, T, A \rangle$, where
\begin{itemize}
\item $C  =  \cup C_i$
is a finite non-empty set of classes describing the subject domain
concepts;
\item $R  =  \cup R_i$
is a finite set of binary relations on classes (concepts), and
$F_R: C\times C\rightarrow 2^R$ is a function defining the names
of binary relations between the classes;
\item $T = \cup T_i$ is a set of data type
with the domain of possible values $\{v_1, \ldots ,v_i\}$;
\item $A = \cup a_i$
is a finite set of attributes, $AK\subseteq  A$ is a subset of the
key attributes for unique identification of the instances of
concepts and relations, and $F_A: C\cup R\rightarrow 2^{A\times
T}$ is a function defining the names and the types of attributes
for classes $C$ and relations $R$.
\end{itemize}
The \emph{information content} of an ontology $O$ is represented
as $IC = \langle I,  RI \rangle$, where
\begin{itemize}
\item $I = \cup I_i$
is a finite set of ontology class instances from $O$ where $I_i$
from class $C_i\in C$ consists of a set of attributes $a_j$ with
values $v_j$: $I_i = \cup_j (a_j, v_j)$ and $(a_j, v_j)\in
F_A(C_i)$;
\item $RI = \cup RI_i$ is ontology relation instances which is
a finite set of relations on the set $I$ of class instances.
Relation instance $RI_i$ of the relation $R_i$ consists of a
instances $(o_1)_i,(o_2)_i\in I$ of classes $C_1$ and $C_2$
respectively, with a set of attributes $a_j$ provided with values
$v_j$: $RI_i = ((o_1,o_2)_i, \cup (a_j, v_j))$, where $R_i\in
F_R(C_1,C_2)$ and $(a_j, v_j)\in F_A(R_i)$.
\end{itemize}

Rules for completing the ontology recognize instances of ontology
classes or relations in input data, evaluate their attributes and
bind class instances in relation instances. A semantic-syntactic
models of input data languages are usually manifold and
complicate. A universal formalization for this topic is out of
scope of this paper.

The preliminary phase of data processing is executed by an
external analysis module based on a vocabulary of the subject
domain. This module constructs (1) a set of \emph{instance} agents
corresponding to ontology concepts, and (2) a set of
\emph{relation} agents corresponding to ontology relations. The
information agents make use of knowledge concerning their
\emph{positions in input data}. This knowledge is represented as a
set of closed natural intervals. We consider this set as set of
natural numbers in sense that two intersecting intervals are
joined into one.

The \emph{rule} agents implement rules of input data processing
and ontology population. According to information received from
instance and relation agents, they generate new attribute values
of the instances and relations, send the obtained result to all
agents interested in it, or generate new instance or relation
agents. Eventually, the information agents assign values to all
their attributes that can be evaluated with the information from
the data, and the system stops. A \emph{controller} agent keeps
track of system stopping. At the termination moment, the instance
agents have accumulated all possible values for each of their
attributes to resolve information ambiguities. Formal definitions
of agents follow.

A set of \emph{instance agents} $IA$ corresponds to class
ontological instances from $O$. Each $I \in IA$ is a tuple $I=(id;
Cl; Atr; Rul; Pos; Rel),$ where
\begin{itemize}
    \item $id$ is a unique agent identifier;
    \item $Cl\in C$ is an ontological class of the agent;
    \item $Atr=\bigcup_{j\in[1..k]}(a_j, V_j, Rul_j, pos_j)$ is a set
    of attributes of the agent, where for each $j\in[1..k]$
    \begin{itemize}
        \item $a_j$ is a name of the agent attribute;
        \item attribute values from $V_j$ belongs to the domain of the corresponding type
         and $(a_j,V_j)\subseteq F_A(Cl_O)$;
        \item every rule agent in set of rule agents $Rul_j$ requires the value of attribute $a_j$ to get the result;
        \item $pos_j$ is a set of closed natural intervals corresponding
        to the attribute position in the input data;
    \end{itemize}
    \item $Rul$ is a set of rule agents that use data included in this instance agent
    as an argument;
    \item $Pos = \bigcup_{j\in[1..k]}pos_j$
    is a set of natural intervals corresponding  to the agent position in the input data;
    \item $Rel$ is a set of possible relations of the agent;
    for every $(r,ir)\in Rel$: $ir$ is a set of instance identifiers of relation
agent $r$ which include this agent.
\end{itemize}

A set of \emph{relation agents} $RlA$ corresponds to ontological
relations from $O$. Each $Rl \in RlA$ is a tuple $Rl=(id; R_O; IR;
Rul; Pos),$ where
\begin{itemize}
    \item $id$ is a unique agent identifier;
    \item $R_O\in R$ is an ontological relation of the agent;
    \item $IR=\bigcup_{i\in[1..k]}((o_1,o_2)_i, Atr_i, pos_i)$ is a set
    of instances of relation $R_O$,
    where for each $i\in[1..k]$
    \begin{itemize}
        \item relation objects $o_1$ and $o_2$ are identifiers of instance
        agents belonging to ontological classes $C_1$ and $C_2$ respectively and $Rl_O\in F_R(C_1,C_2)$;
        \item every relation attribute $(a, v, pos)\in Atr_i$ with name $a$ has
        attribute value $v$ with $(a,v)\in F_A(Rl_O)$ and data position $pos$;
        \item $pos_i =( Pos_{o_1}\cup Pos_{o_2})\bigcup(\cup_{(a, v, pos)\in Atr_i}pos)$
    is a set of natural intervals corresponding to the agent position in the input data;
    \end{itemize}
    relation instance is evaluated iff both its objects are evaluated;
    \item $Rul$ is a set of rule agents that use this relation agent as an argument;
    \item $Pos = \bigcup_{i\in[1..k]}pos_i$
    is a set of natural intervals corresponding to the agent position in the input
    data.
\end{itemize}

A set of \emph{rule agents} $RA$ corresponds to rules of input
data processing and ontology population rules. Each \emph{rule
agent} $R \in RA$ is a tuple $R = (id; Args; make\_res(args),
result)$, where
\begin{itemize}
    \item $id$ is a unique agent identifier;
    \item $Args=\cup(arg_1 (Cl_1),...,arg_s (Cl_s))$ is a set of argument vectors,
    where for each $i\in [1..s]$: $arg_i$ is an argument value
    determined by the corresponding instance or
relation agents from ontological class $Cl_i$; let us denote
vector of arguments' values as $args$, where each value is
\begin{itemize}
    \item an
attribute value provided with the identifier of an instance agent,
    \item an identifier of an instance agent,
    \item an identifier of an instance of a relation agent;
\end{itemize}
    \item $make\_res(args)$ is a function computing the result
    from vector $args$;
    \item $result$ is the result of function $make\_res(arg)$
    which can be
    \begin{itemize}
        \item empty, if the argument vector is inconsistent;
        \item values of some attributes with their positions for some instance agents and/or
        \item tuples of values of some objects and attributes with their positions for some relation agents and/or
        \item new information agents (they must differ from other agents by their classes and values of attributes).
    \end{itemize}
\end{itemize}

As a simple example let us consider the following multi-agent
system for natural language text processing. Let the given
ontology includes classes $SciEvent$, $GeoPlace$, and relation
$Venue$. The corresponding instance and relation agents have the
following form:
\begin{itemize}
\item[$\bullet$]
 $I_1 = (\emptyset; SciEvent;
\\\hspace*{\fill}
 (date, \{R\_Calendar, \ldots\},\emptyset(Dates),\emptyset),
 (name, \{R\_Calendar, R\_Person,\ldots\},\emptyset(String),\emptyset),\ldots;
\\\hspace*{\fill}
 \{R\_Venue, R\_Date, R\_Person,\ldots\};\emptyset; \{(Venue,\emptyset), (OrganizedBy,\emptyset),\ldots\})$.
\\ $SciEvent$ has attributes $date$ and $name$ which can be used by
rule agents $R\_Calendar$, $R\_Person$ and others. The agent
itself is used by rule agents $R\_Venue$, $R\_Date$, $R\_Person$
and others. The relations of the agent are $Venue$, $OrganizedBy$
and others.
\item[$\bullet$]
 $I_2 = (\emptyset; GeoPlace;
\\ (name, \{R\_Venue, R\_GeoPlace,\ldots\},\emptyset(String),\emptyset),
\\\hspace*{\fill} (country, \{R\_GeoPlace, \ldots\},\emptyset(Countries),\emptyset),\ldots;\hspace*{\fill}
\\\hspace*{\fill}
 \{R\_Venue, R\_Travel, \ldots\};\emptyset; \{(Venue,\emptyset), (BirthPlace,\emptyset),\ldots\})$.
\\ $GeoPlace$ has attributes $name$ and $country$ which can be used by
the corresponding rule agents $R\_Venue$, $R\_GeoPlace$ and
others. The agent itself is used by rule agents $R\_Venue$,
$R\_Travel$, and others. The relations of the agent are $Venue$,
$BirthPlace$ etc.
\item[$\bullet$]
 $Rl_1 = (1; Venue; IR((SciEvent,GeoPlace),\emptyset); \emptyset;\emptyset)$.
\\ $Venue$ connects scientific events and geographic places.
\end{itemize}

As an example of a rule agent let us consider agent $R\_Venue$:
\\ $R\_Venue = (1, arg_1(SciEvent), arg_2(GeoPlace), arg_3(HoldOp);
 \\\hspace*{1.5em} (1)\ Caption(\{arg_1,arg_2\}), Preposition(arg_1,arg_2)\ ||
 \\\hspace*{1.5em} (2)\ Reference(\{arg_1,arg_2\}), Preposition(arg_1,arg_2)\ ||
 \\\hspace*{1.5em} (3)\ Sentence1(\{arg_1,arg_2\}), Preposition(arg_1,arg_2),
 \\\hspace*{\fill} BracketSegment(\{arg_2\}), Contact\_Stop(arg_1,arg_2)\ ||
 \\\hspace*{1.5em} (4)\ Sentence2(\{arg_1,arg_2,arg_3\}),
 \\\hspace*{\fill} Preposition(arg_1,arg_3), Contact\_NegWords(arg_1,arg_2), \hspace*{\fill}
 \\\hspace*{\fill} Preposition(arg_3,arg_2), Contact\_Attr(arg_3,arg_2,arg1)\ ||
 \\\hspace*{1.5em} (5) \ldots;
\\\hspace*{1.5em} Venue.o_1=arg_1, Venue.o_2=arg_2)$.
\\ This rule agent matches scientific events to geographic places.
It can recognize this matching in captions, references, various
sentences taking into account mutual positions of its arguments
and their contacts (for instance, events and places can be
interpointed in $Sentence1$). Third argument $arg_3(HoldOp)$
accumulates all verbs and phrases indicating venue: `hold',
`locate', `take place' etc. Let us consider the following part of
MOD* call for papers:
\begin{quote}
The 1st Workshop on Logics and Model-Checking for Self-* Systems
(MOD*)
\\ http://modstar.cs.unibo.it/
\\ 12 September 2014, Bertinoro, Italy
\end{quote}
The following evaluation of attributes of the above agents is the
result of analysis of the given text fragment:
\begin{itemize}
    \item  $I_1 = (1;\ SciEvent;\  (date, \{\ldots\}, 12.09.2014, [13,15]),
\\\hspace*{\fill} (name, \{...\},
 \mbox{"The 1st Workshop on Logics and Model-Checking for Self-* Systems"},[1,10]),...;
\\\hspace*{\fill}
 \{\ldots\};\ \{[1,10],[13,15]\};\ \{(Venue,1), (OrganizedBy,\emptyset),\ldots\})$.
    \item  $I_2 = (2;\ GeoPlace;\
 (name, \{\ldots\},\mbox{Bertinoro},[16]),\
 (country, \{\ldots\},Italy,[17]),\ldots;
\\\hspace*{\fill}
 \{\ldots\};\ [16,17];\ \{(Venue,1), (BirthPlace,\emptyset),\ldots\})$.
    \item  $Rl_1 = (1;\ Venue;\ \{(1,\ 2,\ \{[1,10],[13,17]\})_1\};\ \emptyset;\ \{[1,10],[13,17]\})$.
\end{itemize}

Now we give brief overview of interactions of the above
information and rule agents. Multi-agent system $\mathbf{MDA}$ for
data analysis includes information agents sets, a rule agents set,
and an agent-controller. The result of agent interactions by
protocols below is data analysis, when the information agents
determine the possible values of their attributes and objects from
a given data. All agents execute their protocols in parallel. That
is, all agents act in parallel until none of the rule agent can
proceed. These termination event is determined by the controller
agent. We use an original algorithm for termination detection
which is based on activity counting. The system is dynamic because
rule agents can create new information agents.

The agents are connected by duplex channels. The controller agent
is connec\-ted with all agents, instance agents are connected with
their relation agents from $Rel$, and information agents are
connected with rule agents that use information from them and/or
provide new attribute/object values for them. Messages are
transmit\-ted instantly in a reliable medium and stored in
channels until being read.

Let $IA = \{I_1, ..., I_n,...\}$ be an instance agents set, $RlA =
\{Rl_1, ..., Rl_m,...\}$ be a relation agents set, and $RA =
\{R_1, ..., R_s\}$, be a rule agents set. The result of executing
of the following algorithm is data analysis, when the information
agents determine the possible values of their attributes. Let
\texttt{Ii} be a protocol of actions of instance agent $I_i$,
\texttt{Rlj} be a protocol of actions of relation agent $Rl_j$,
and \texttt{Rk}, be the protocol of actions of rule agent $R_k$,
\texttt{C} be the protocol of actions of an agent-controller $C$.
Then the multi-agent data analysis algorithm $\mathbf{MDA}$ can be
presented in pseudocode as follows:
\\ \texttt{MDA::
\\\hspace*{1em}  parallel   \{I1\} \ldots \{In\} \ldots \{Rl1\} \ldots \{Rlm\} \ldots \{R1\} \ldots \{Rs\} \{C\}
}

Here the \texttt{parallel} operator means that all execution flows
(threads) in the set of braces are working in parallel. Brief
descriptions of the protocols follow.

Let further $C$ be the controller agent; $R, R_{ij}$ be rule
agents; $I$ be an instance agent; $Rl$ be a relation agent; $A$ be
an information agent; \texttt{mess} be message (special for every
kind of agents); $Input$ be queue of incoming messages. We suppose
that all specialities are clear from the context. For the
simplicity, we suggest that rule agents produce results with at
most one attribute per an instance agent and/or at most one
instance of relation per a relation agent. This case could be
easily generalized for multiple results.

Informal description of the instance agent protocol. In the first
phase of its activities the instance agent sends evaluated data to
all rule agents interested in these data. Then the agent processes
the received data by updating its attributes, relations, and
increasing the position with the attributes' positions, sending fresh
attribute values to rule agents interested in. Every change of
activity is reported to the controller agent. The instance agent
terminates if it receives the stop message from the controller agent.

\noindent \textbf{Protocol of instance agents.}
\\ \texttt{ $I$::
\\ 1.\hspace*{0.6em} send $|Rul|+1$ to $C$;
\\ 2.\hspace*{0.6em} forall $R \in Rul$ send $id$ to $R$;
\\ 3.\hspace*{0.6em} forall $a_{i}\in Atr$
\\ 4.\hspace*{2.2em}        if $a_i\neq\emptyset$ then \{ send $|Rul_{i}|$ to $C$;
\\ 5.\hspace*{3.6em}            forall $R_{ij} \in Rul_{i}$ send $a_i$ to $R_{ij}$;\}
\\ 6.\hspace*{0.6em} send $-1$ to $C$;
\\ 7.\hspace*{0.6em} while (true)\{
\\ 8.\hspace*{1.2em}      if $Input\neq\emptyset$ then \{
\\ 9.\hspace*{2.8em}        mess = get\_head($Input$);
\\ 10.\hspace*{2.2em}        if mess.name = $C$ then break;
\\ 11.\hspace*{2.2em}        if mess.name $\in Rel$ then upd\_Rel(mess.name, mess.id);
\\ 12.\hspace*{2.2em}        if mess.id = $i$ then \{
\\ 13.\hspace*{3.7em}             upd($a_{i}$, mess.value, mess.pos);
\\ 14.\hspace*{3.7em}             upd($Pos$, mess.pos);
\\ 15.\hspace*{3.7em}             send $|Rul_{i}|$ to $C$;
\\ 16.\hspace*{3.7em}             forall $R_{ij} \in Rul_{i}$ send $a_{i}$ to $R_{ij}$; \}
\\ 17.\hspace*{2.2em}        send $-1$ to $C$; \} \}
}

Informal description of the relation agent protocol. In the first
phase of its activities the relation agent sends evaluated data to
all instance and rule agents interested in these data. The agent
processes the received data by updating instances of its objects,
attributes and increasing the position with the objects' and
attributes' positions, sending identifiers of these fresh
instances to instance agents included into evaluated tuples of
data. Every change of activity is reported to the controller
agent. The relation agent terminates if it receives the stop
message from the controller agent.

\noindent \textbf{Protocol of relation agents.}
\\ \texttt{ $Rl$::
\\ 1.\hspace*{0.6em} send $1$ to $C$;
\\ 2.\hspace*{0.6em} forall $ir_i\in IR$
\\ 3.\hspace*{1.6em}   if evaluated($ir_i$) then \{
\\ 4.\hspace*{3.1em}      send $|Rul|+2$ to $C$;
\\ 5.\hspace*{3.1em}      send ($Rl, ir_i$) to $(o_1)_i$ and $(o_2)_i$;
\\ 6.\hspace*{3.1em}      forall $R \in Rul$ send ($Rl, ir_i$) to $R$;\}
\\ 7.\hspace*{0.6em} send $-1$ to $C$;
\\ 8.\hspace*{0.6em} while (true)\{
\\ 9.\hspace*{1.6em}      if $Input\neq\emptyset$ then \{
\\ 10.\hspace*{2.8em}          mess = get\_head($Input$);
\\ 11.\hspace*{2.8em}          if mess.name = $C$ then break;
\\ 12.\hspace*{2.8em}          upd\_Rel(mess.id, mess.value, mess.pos);
\\ 13.\hspace*{2.8em}          $i$ = mess.id
\\ 14.\hspace*{2.8em}          if evaluated($ir_i$) then \{
\\ 15.\hspace*{4.3em}                send $|Rul|+2$ to $C$;
\\ 16.\hspace*{4.3em}                send ($Rl, ir_i$) to $(o_1)_i$ and $(o_2)_i$;
\\ 17.\hspace*{4.3em}                forall $R \in Rul$ send ($Rl, i$) to $R$;\}
\\ 18.\hspace*{4.2em}          send $-1$ to $C$; \}\}
}

Informal description of the rule agent protocol. It has two parallel
sub\-pro\-ces\-ses: processing incoming data from instance agents
(\texttt{ProcInput}) and producing the outcoming result
(\texttt{ProcResult}). Processing incoming data includes (1)
for\-ming argument vectors, and (2) sending argument vectors or
indication of termi\-nation to \texttt{ProcResult}. Producing the
outcoming result includes (1) checking conformity of arguments and
argument vectors, (2) making the result, which is new attribute
values of some information agents and/or new information agents with
their positions, and (3) de\-ter\-mining agents for sending new
values to. New information agents start im\-me\-dia\-te\-ly with data
given them by the rule agent at birth. Every change of activity is
reported to the controller agent. The rule agent terminates if it
receives the stop message from the controller agent.

\noindent \textbf{Protocol of rule agents.}
\\ \texttt{$R$ ::
\\\hspace*{0.3em} SendList: set of Instance Agents = \mbox{$\emptyset$};
\\ 1.\hspace*{0.3em}  parallel
\\ 2.\hspace*{0.6em}  \{  ProcInput$_R$; ProcResult$_R$; \}
\\ ProcInput$_R$ ::
\\\hspace*{0.3em} $args$: set of vectors of Argument;
\\ 1.\hspace*{0.6em}  while (true) \{
\\ 2.\hspace*{2em}       if $Input\neq\emptyset$ then \{
\\ 3.\hspace*{3.2em}          mess = get\_head($Input$);
\\ 4.\hspace*{3.2em}          if mess.name=$C$ then \{
\\ 5.\hspace*{4.65em}                send `stop' to ProcResult$_R$;
\\ 6.\hspace*{4.65em}                break; \}
\\ 7.\hspace*{3.2em}          if mess.name=$A$ then \{
\\ 8.\hspace*{4.65em}               $args$ = make\_arg(mess.value, $A$);
\\ 9.\hspace*{4.65em}                  if ($args\neq\emptyset$) send ( $args$ ) to ProcResult$_R$;
\\ 10.\hspace*{4.2em}               send $|args|-1$ to $C$; \}\}\}
\\ ProcResult$_R$ ::
\\\hspace*{0.3em} $arg$: vector of Argument$\cup$\{`stop'\};
\\ 1.\hspace*{0.6em}  while (true) \{
\\ 2.\hspace*{2em}      if $Input \neq\emptyset$ then \{
\\ 3.\hspace*{3.2em}          $arg$ = get\_head($Input$);
\\ 4.\hspace*{3.2em}          if $arg$ = `stop' then break;
\\ 5.\hspace*{3.2em}          $(result,SendList)$ = make\_res($arg$);
\\ 6.\hspace*{3.2em}          if $result\neq\emptyset$ then \{
\\ 7.\hspace*{5em}                  start\_new\_information\_agents;
\\ 8.\hspace*{5em}                  send $|SendList|$ to $C$;
\\ 9.\hspace*{5em}                  forall $A \in SendList$ send $result(A)$ to $A$;\}
\\ 10.\hspace*{2.7em}          send $-1$ to $C$; \}\}
}

The main job of the controller agent is to sequentially calculate
other agents' activities. If all agents are inactive, the agent sends
them all the stop message.

\noindent \textbf{Protocol of agent-controller $C$.}
\\ \texttt{ C ::
\\\hspace*{0.3em} $Act$: integer;
\\\hspace*{0.3em} {\it Input}: set of integer;
\\ 1.\hspace*{0.5em}  $Act$ = 0;
\\ 2.\hspace*{0.5em}  while( ${\it Input}=\emptyset$ ) \{  \}
\\ 3.\hspace*{0.5em}  while(true)\{
\\ 4.\hspace*{1.2em}      if( ${\it Input}\neq\emptyset$ ) then $Act$ = $Act$ + get\_mess({\it Input});
\\ 5.\hspace*{1.2em}      if( ${\it Input}=\emptyset$ and $Act$ = 0 ) then break; \}
\\ 6.\hspace*{0.5em}   send STOP to all;
}

The following proposition is straight consequence of Proposition 1
from \cite{GaraninaSidorovaBodin13}:
\begin{proposition}\label{MDAmain}
Multi-agent system $\mathbf{MDA}$ terminates and the
agent-controller determines the termination moment correctly.
\end{proposition}
The proposition is proved in \cite{GaraninaSidorovaBodin13}. The
proof of the first part is based on finiteness of input data and
reasonable suggestion that rules of ontology population and data
processing cannot generate new information infinitely. The second
assertion is based on timely notifying about activities of
information and rule agents.

\section{Model Checking of Multi-agent Data Analysis}\label{APPR}

We would like to verify properties from proposition \ref{MDAmain}
formally because a parallel interaction of agents is rather knotty.
The crucial property of this multi-agent system is termination.
Another important property is correctness of actions of the
agent-controller, i.e. that the agent correctly detects the moment of
system termination when all system agents do nothing just waiting
messages from others. Besides, there is an interesting
``operability'' property: in a future at least one information agent
will update at least one of its attributes. Specific properties of
soundness and completeness of information processing are also very
important, but we think it is practically impossible to check these
by formal verification methods.

Our agent model is finite if we suggest that rules of ontology
population and data processing do not generate new information
agents infinitely. Hence it is possible to use finite-state model
checking technique for verification. For this it is reasonable to
code the model in integers. Let us explain the approach by an
example of semantic text analysis for ontology population.

\noindent \textbf{(1) Input data.} As input data we have a finite
natural language text, hence we can just enumerate words in this
text.

\noindent \textbf{(2) An ontology.} We suggest that a given
ontology has a finite number of classes and relations and
attribute values of classes and relations belong to finite domains
or input data\footnote{For example, let input data be texts of
calls for papers for conferences, then important dates of a
conference can be an attribute of class $Conference$ in ontology
$Scientific Events$.}. Hence we can enumerate instances of classes
and relations and their attributes.

\noindent \textbf{(3) A model of the domain-specific language.} A
special preprocessing module based on this model constructs finite
number of information agents corresponding to input text. Every
information agent (its attributes) can contain the following
descriptive information:
\begin{itemize}
    \item ontological: belonging to numerated classes or
    relations, holding numerated evaluated values of some attributes;
    \item grammatical: enumerated morphological and syntactical characters;
    \item structural: an enumerated text position;
    \item segmental: belonging to an enumerated text segment\footnote{For example, to ``Conference Topics''
    in calls for papers for conferences.}.
\end{itemize}
Again we can enumerate these agents and their inside information.

\noindent \textbf{(4) Processing rules.} Every processing rule
agent uses a set of arguments whose values come from information
agents and include all necessary details (ontological,
grammatical, structural and segmental). These data are represented
by natural numbers. A rule agent produces result as a set of
natural numbers forming attribute values or new information
agents. These values and elements of the new agents are some
arguments with descriptive information or they belong to
corresponding domains. Hence, rule agents consume integer numbers
and produce integer numbers.

A reason for our system termination is that position $Pos$ of
every information agent cannot increase infinitely since it is
bounded by number of words in the input text. Besides, we have to
be sure that there is no infinite information for these positions.
This property can be formulated using a special construct
according to every vector of rule arguments. Let position
$Pos(arg)$ be a union of positions of arguments from $arg$. Let
call it a \emph{position point}. Informally, this position point
corresponds to a set of words from the input text located in
positions from $Pos$. It is reasonable to limit amount of new
information corresponding to one position point. Hence, in order
to verify the property of interest for every rule agent we just
have to accumulate numbers of new information items for every
position point of the rule, and then compare them with the limit.
This limit depends on the degree of \emph{terminological homonymy}
of the domain-specific language. We say that two sets of words are
terminological homonyms iff they include vocabulary terms which
are homonyms. This fact causes generation of several ontology
objects simultaneously associated with these sets. This homonymic
limit $HomLim$ is the same for every rule. Now the termination
property can be formulated as follows: if rule agents can not
produce information more than the homonymic limit then the system
stops.

\section{Using SPIN for model checking $\mathbf{MDA}$}\label{SPIN}

For formal model checking of our multi-agent data analysis system
we choose popular and well-develo\-ped model checking tool
SPIN\cite{Holzmann03}. We have tried NuSMV model checker also, but
have found that its input language is not fit to our multi-agent
model because a lot of arrays in the model make the corresponding
NuSMV-model very complicated. For verification SPIN requires a
model of the system written on SPIN input language Promela with
model properties expressed in linear time logic LTL.

SPIN deals with finite data only. The previous section justifies
the following simplification of the original model of data
analysis: (1) as input data for analysis we consider finite sets
of integers from a bounded integer interval; (2) attribute values
of class and relation instances of an ontology are integers; (3)
the result of rule agent actions is tuples of integers as
attribute values for information agents. Thus in this simplified
model of protocols above it does not matter what exact values are
processed by our agents. We are only interested in verification of
termination, operability and correctness of termination detection.

For the Promela model specification we define agent processes
$InstAgent$, $RelAgent$, $RulAgent$ and $Controller$ corresponding
to agents of the model above. Agents are instances of processes of
the corresponding type. $Controller$ is the main process which run
all other processes at the beginning. SPIN assigns unique
identification number $\_pid$ to every process. Further we
describe some features of these processes.

\textbf{(1) Structures of processes}. These structures are based
on definitions of agents from \ref{AgentModel}. They include
Promela structures with fields con\-tai\-ning integer arrays. For
example, the following code is a part of an instance agent
definition:
\begin{verbatim}
proctype ins_agent(){
    byte id;
    d_step{
        INS_AGENT[INS_AGENT_COUNT] = _pid;
        INS_AGENT_COUNT = INS_AGENT_COUNT + 1;
        id = INS_AGENT_COUNT;           // unique agent identifier
    }

    int Class;                          // class of the agent
    int RuleOut[MAX_RULE_OUT];          // rules Rul
    Attribute attrs[MAX_ATTR];          // attributes of the agent
    Relation Relations[NUM_INS];        // relations of the agent

    MessagetoRule toRule;
...
\end{verbatim}

\textbf{(2) Types of communicating messages.} They are different
for different process types and also implemented as Promela
structures with fields containing integers and integer arrays. The
following code demonstrates messages to an instance agent and to a
rule agent:
\begin{verbatim}
typedef MessagetoIns{
 int name;          // name of the sender
 int id;            // name of the relation instance (if any)
 int vals_id;       // name of the attribute (if any)
 int vals_value;    // value of the attribute (if any)
}

typedef MessagetoRule{
 mtype type;        // { Agent, Attribute, Relation }
 int name;          // name of the sender
 int val;           // value of the attribute or name of the relation instance
}
...
\end{verbatim}

\textbf{(3) Agents initialization.} We assign initial data to
information agents which imitates a result of work of the external
module for preliminary data analysis. We implement this
initialization depending on number $\_pid$. This number defines a
class of an agent, its outcoming rules $Rul$ and $Rul_i$ for every
attribute $a_i$ (see the definition of information agents), and
its evaluated attributes. The following code is the example of an
instance agent initialization:
\begin{verbatim}
 Active = 0;                                    // agent activity
 Class = id;
 for (i : 0 .. id-1 ) { RuleOut[i] = i+1; }     // rules Rul
 for (i : 0 .. 2*id-1) {
    attrs[i].RuleOut[0] = i/2+1;                // rules of attributes
    if
        :: (i%2 == 0) ->
            attrs[i].values[0] = i/2+1;         // values of attributes
            attrs[i].values_count = 1;
        :: else -> skip;
    fi;
 }
 for (i : 0 .. id-1 ) { Relations[i].name = i+1; }  // relations
...
\end{verbatim}

\textbf{(4) Agent actions.} They are based on the protocols above
and include message passing and agents' data updating. Rule agents
also create argument vectors and compute the result for
information agents which models rules of data processing and
ontology population. Actions of information agents and the
controller can be translated to Promela almost directly from
protocols of the previous section. Subprocesses of each rule agent
correspond to consuming input information and producing output
information. Function $make\_arg$ of a rule agent defines a
position of incoming data in a vector of arguments with respect to
the rule agent definition. This function forms the next data (a
set of argument vectors) for processing in function $make\_res$
that imitates using rules of forming results by input data that
depend on the structure of these data and the ontology. Here this
imitation depends on input argument values and $\_pid$ of the
rule-process. These parameters are used to define: (1) consistency
of an argument vector; (2) quantity and numbers of instance
agents, their attributes, that must be updated and values of these
attributes; (3) quantity and numbers of relation agents, their
instances that must be updated or added, elements of the relation
to be changed and their new values. All imitation function are
very simple because they are just intended to simulate a linear
computation time of real functions for making results to send.
Note that function $make\_arg$ has an exponential time complexity.
The following code simulates generating a new attribute value for
an instance agent:
\begin{verbatim}
proctype rule_agent(){               // start consuming subprocess
    byte id;                         // unique agent identifier
    ... ...
    run rule_agent_out(_pid, id);    // start producing subprocess instance
 ... ...
   for(i: id .. 2*id-1){
    mti.vals_id = id+1;                     // name of the updated attribute
    mti.vals_value = argV[i].val;           // new attribute value
    toController ! ( 1 );                   // info for controller
    toInsAgent[ argV[i].name ] ! ( mti );   // send the new value
   }
 ... ...
}
\end{verbatim}

Let us express properties of the model to be verified. Let every
agent $A\in IA\cup RlA\cup RA$ (not the controller) have a special
boolean activity status $A.active$, whose value is $true$ when the
agent does something useful (sends or processes messages) and $false$
when the agent just waits for messages and does nothing else. Thus
the controller correctness property can be expressed in LTL as
$$G(Act=0 \rightarrow \bigwedge_{A\in IA\cup RlA\cup
RA}A.active=false)$$. Initially its value is $false$ and after the
first message it becomes $true$. The operability property can be
expressed as
$$F(\bigvee_{I\in
IA}I.was\_upd=true)\vee(\bigvee_{Rl\in RlA}Rl.was\_upd=true)),$$
where $A.was\_upd$ is a boolean variable recording that agent $A$
has updated its attribute, i.e. initially it is set to $false$ and
after the first attribute update it becomes $true$.

The termination property is expressed in LTL as
$$G(\bigwedge_{R\in RA}R.gen < R.pnt \cdot HomLim) \rightarrow F(\bigwedge_{A\in IA\cup
RlA\cup RA}A.active=false),$$ where $R.gen$ is the number of new
information agents generated by rule agent $R$, $R.pnt$ is the
number of position points corresponding to these generations.

\section{Conclusion}\label{CONC}

In the paper we suggest the approach to verification of the
multi-agent data analysis algorithm for ontology population. A
means of verification is model checking tool SPIN and properties
of the system are expressed by LTL-formulas. We simulate our model
in SPIN for fifty main agents and the agent-controller (138130
steps). SPIN appears to be able to make exhaustive verification of
termination correctness and operability properties for twelve main
agents, and bitstate verification for eighteen main agents. The
latter verification required 25~minutes of a computer with an
Intel Celeron(R) CPU running at 2.6~GHz and about 1~GByte of RAM.
Both properties are satisfied in this model. The termination
property have not verified yet.

At this stage of our research we do not handle competition and
cooperation of instance agents for resolving ambiguities. We plan
to enrich the agents with abilities for these kinds of
interactions, to develop ambiguity resolving algorithms and to
verify their properties such as termination, correctness of
interactions etc.

\nocite{*}
\bibliographystyle{eptcs}
\bibliography{GaraninaMOD}

\end{document}